\documentclass[10pt, a4paper]{article}

\usepackage{lrec-coling2024} 

\usepackage{multirow}
\usepackage{booktabs}
\usepackage{graphicx}
\usepackage{latexsym}
\usepackage{caption}
\usepackage{subcaption}
\usepackage{color}
\usepackage{soul}
\usepackage{graphicx}
\usepackage{svg}
\usepackage{textcomp}
\usepackage{url}
\usepackage{siunitx}
\usepackage{gensymb}
\usepackage{amsmath}
\usepackage{amssymb}
\usepackage{pifont}

\definecolor{MyCyan}{HTML}{95FFF6}

\usepackage[normalem]{ulem} 
\usepackage{algorithm}
\usepackage{algpseudocode}
\makeatletter
\def\BState{\State\hskip-\ALG@thistlm}
\makeatother

\usepackage{physics} 

\title{Evaluating Document Simplification: On the Importance of Separately Assessing Simplicity and Meaning Preservation}

\name{Liam Cripwell$^\dagger$, Joël Legrand$^{\dagger,\ddagger}$, Claire Gardent$^\dagger$} 

\address{$^\dagger$LORIA, CNRS, Inria, Université de Lorraine, Nancy, France \\
        $^\ddagger$Centrale Supélec, Metz, France \\
        \{liam.cripwell, joel.legrand, claire.gardent\}@loria.fr\\}

\abstract{
Text simplification intends to make a text easier to read while preserving its core meaning. Intuitively and as shown in previous works, these two dimensions (simplification and meaning preservation) are often-times inversely correlated. An overly conservative text will fail to simplify sufficiently, whereas extreme simplification will degrade meaning preservation. Yet, popular evaluation metrics either aggregate meaning preservation and simplification into a single score (SARI, LENS), or target meaning preservation alone (BERTScore, QuestEval). Moreover, these metrics usually require a set of references and most previous work has only focused on sentence-level simplification. In this paper, we focus on the evaluation of document-level text simplification and compare existing models using distinct metrics for meaning preservation and simplification. We leverage existing metrics from similar tasks and introduce a reference-less metric variant for simplicity, showing that models are mostly biased towards either simplification or meaning preservation, seldom performing well on both dimensions. Making use of the fact that the metrics we use are all reference-less, we also investigate the performance of existing models when applied to unseen data (where reference simplifications are unavailable).
 \\ \newline \Keywords{simplification, evaluation, out-of-domain} }

\begin{document}

\maketitleabstract

\section{Introduction}
Text simplification is the task of rewriting a text such that it is easier read and understood by a wider audience, while still conveying the same central meaning. This generally involves transformations such as lexical substitution~\citep{wrro132086,north2023deep} or structural modifications (sentence splitting) according to the text syntax~\citep{narayan-etal-2017-split} or discourse structure~\citep{niklaus-etal-2019-transforming,cripwell-etal-2021-discourse-based}. Although the main motivation is to promote accessibility~\citep{williams-etal-2003-experiments,kajiwara-etal-2013-selecting}, it can also be a useful preprocessing step for downstream NLP systems~\citep{miwa-etal-2010-entity,mishra-etal-2014-exploring,stajner-popovic-2016-text,niklaus-etal-2016-sentence}. 

While early simplification work has focused on individual sentence inputs~\citep{nisioi-etal-2017-exploring,martin-etal-2020-controllable,cripwell-etal-2022-controllable,yanamoto-etal-2022-controllable}, recent progress has been made on document-level simplification~\citep{sun-etal-2021-document,cripwell-etal-2023-document,cripwell-etal-2023-context}. However, several challenges stand in the way of further progress on simplification tasks, including the limited ability to transparently perform automatic evaluation and most popular metrics' requirement of multiple references.

Recent investigation into the quality of sentence-level test data and system outputs has found many instances of factual incoherence not previously detected during data collection or evaluation~\citep{devaraj-etal-2022-evaluating}. This raises questions of how faithful simplifications are to their inputs and whether or not these concerns also apply to the document-level task. Although attempts to automatically evaluate semantic faithfulness in sentence simplification have seen limited success~\citep{devaraj-etal-2022-evaluating}, summarization literature contains a lot of work that could be transferable to document simplification~\citep{laban-etal-2022-summac,fabbri-etal-2022-qafacteval}.

Despite their ability to generate highly fluent texts, the commonly used end-to-end neural systems rely heavily on the quality of data they are trained on. In text simplification, training data is scarce, with most existing corpora being compiled via automatic alignment methods. These are known to contain a lot of noise and imbalanced distributions of possible transformation types~\citep{sulem-etal-2018-simple,jiang-etal-2020-neural}. As a result, end-to-end systems are very conservative in the amount of editing they perform, often making little to no changes to the input~\citep{alva-manchego-etal-2017-learning}. With some works observing an inverse correlation between meaning preservation/faithfulness and simplicity~\citep{schwarzer2018human,vu-etal-2018-sentence}, this raises the question of whether those models sufficiently simplify the input text (since some amount of degradation is a requirement for performing simplification).


Evaluation poses additional challenges, with the suitability of popular automatic metrics remaining unclear~\cite{alvam2021Unsuitability,questeval_simp,cripwell-etal-2023-simplicity}. As most automatic metrics require multiple, high-quality references, studies are usually restricted to a small pool of imperfect datasets that include reference simplifications, making it difficult to gauge how well systems actually perform on real-world out-of-domain data. Furthermore, most metrics produce a single score that aims to quantify overall quality, despite the fact that quality aspects are often highly correlated or definitionally at odds with each other~\citep{schwarzer2018human,vu-etal-2018-sentence}. As such, results are often difficult to interpret, making it unclear where models succeed and fail.

In this work, we compare various document-level simplification models in terms of meaning preservation and simplicity, with specific focus on English-language data. Departing from single-value, reference-based scores such as SARI or BERTScore, we exploit distinct, reference-less metrics for these two dimensions. For meaning preservation, we rely on existing reference-less metrics such as SummaC~\citep{laban-etal-2022-summac}, QAFactEval~\citep{fabbri-etal-2022-qafacteval}, entity matching and BLEU~\citep{papineni-etal-2002-bleu} with respect to the input document.

For simplicity, we introduce a variation of the SLE metric proposed in \cite{cripwell-etal-2023-simplicity}, which we refer to as $\epsilon$SLE. It is able to estimate how close a simplification is to the target reading level without relying on any references. We also report the Flesch-Kincaid grade level (\textbf{FKGL})~\citep{kincaid1975derivation}, a simple document readability metric which is based on a regression model that considers the average length of sentences and syllable count of words in the document. 

To assess how well existing models perform on each of these two dimensions, 
we  apply these metrics to the output of four document level simplification models using both in and out of domain test data. We find that none of these four models ranks first on both dimensions, confirming the tension between meaning preservation and simplification. Models with high meaning preservation scores tend to be conservative and under-simplify. Conversely, models that simplify more tend to under-perform in terms of meaning preservation. We further show that for a given model, the trade-off may invert when evaluating on an out-of-domain test set. 



\section{Related Work}

\paragraph{Document Simplification.}
Document simplification work began by iteratively applying sentence simplification methods over documents \citep{Woodsend2011WikiSimpleAS,alva-manchego-etal-2019-cross}, which was quickly found to be insufficient for certain operations, often leading to damaged discourse coherence~\citep{siddharthan-2003-preserving,alva-manchego-etal-2019-cross}. Some works then began reducing the problem scope, focusing on specific subtasks of document simplification, including sentence deletion~\citep{Zhong_Jiang_Xu_Li_2020,zhang-etal-2022-predicting}, insertion~\citep{srikanth-li-2021-elaborative}, and reordering~\citep{lin-etal-2021-towards-document-level}.

\citet{sun-etal-2020-helpfulness} used a sentence-level model with additional encoders to embed tokens from the preceding and following sentences, which they attend to during generation. However, this proved incapable of outperforming a sequence-to-sequence baseline~\citep{sun-etal-2021-document}. \citet{cripwell-etal-2023-document} achieved state-of-the-art performance by first using high-level document context to generate a document plan and then using this plan to guide a sentence simplification model downstream. Later, \citet{cripwell-etal-2023-context} iterated on this framework by exploring the importance of context within the simplification component and proposing several alternate downstream models that lead to further performance increases. 

\paragraph{Faithfulness in Simplification.} The goal of text simplification is not only to make a text easier to read, but also to ensure the same information is conveyed. Until recently, explicit evaluation of the faithfulness of simplification outputs has been somewhat overlooked. In general, semantic adequacy with the original complex text is only manually considered during human evaluation, with automatic metrics mostly focusing on semantic similarity to reference simplifications (which are assumed to be sufficiently faithful). Even during human evaluation, the typical criterion for faithfulness is rather relaxed, demanding only that the text continues to generally convey the core meaning. 

A recent manual investigation into common faithfulness errors in both system outputs and test data found many issues undetected by common evaluation metrics~\citep{devaraj-etal-2022-evaluating}. However, this analysis was limited to sentence-level simplification and many of the issues uncovered do not extend to the document-level case --- a limitation which the authors acknowledge. For instance, content that appears to be wrongly inserted or deleted when considering a pair of aligned sentences in isolation could easily have been moved to or from other sentences in the same document. They also attempted to train a model to automatically evaluate faithfulness, to limited success.
 
Outside of explicit evaluation, some sentence simplification works have considered faithfulness within their training processes. \citet{guo-etal-2018-dynamic} train a multi-task simplification model with entailment as an auxiliary task. \citet{nakamachi-etal-2020-text} integrate the semantic similarity between an input and generated output within the reward function of their reinforcement learning (RL) framework for simplification, while \citet{laban-etal-2021-keep} include an inaccuracy guardrail that rejects generated sequences that contain named entities not present in the input. \citet{ma-etal-2022-improving} attempt to improve performance by down-scaling the training loss of examples with similar entity mismatches. However, these works either do not explicitly evaluate the faithfulness of their system outputs or find that they do not actually prevent the final model from generating unfaithful simplifications.

On the related task of summarization, there has been much more work on this front~\citep{maynez-etal-2020-faithfulness,pagnoni-etal-2021-understanding}. The evaluation of semantic faithfulness in summarization is broadly split into either entailment-based~\citep{falke-etal-2019-ranking,kryscinski-etal-2020-evaluating,koto2022ffci} or question answering (QA)-based methods~\citep{wang-etal-2020-asking,durmus-etal-2020-feqa,scialom2021questeval}, with comprehensive benchmarks being established for each~\citep{laban-etal-2022-summac,fabbri-etal-2022-qafacteval}.

\paragraph{Simplicity Evaluation.}
The most popular evaluation metrics (e.g. SARI, BERTScore) used in simplification generally require multiple high-quality references to perform as intended~\citep{xu-etal-2016-optimizing,zhang2019bertscore}. This poses problems for practitioners seeking to apply simplification models to novel data, as it is impossible to gauge performance without going through the difficult and expensive process of manually creating references ---  a problem that is exacerbated in the document-level case.

Recent investigations into the validity of these metrics also raise concerns over whether they do in fact measure simplicity itself and not correlated attributes like semantic similarity to references~\cite{questeval_simp,cripwell-etal-2023-simplicity}. However, a reference-less sentence simplicity metric (showing high correlations with human judgements) has also been recently proposed, which could allow for meaningful evaluation of out-of-domain performance~\citep{cripwell-etal-2023-simplicity}. Despite this, the efficacy of existing evaluation metrics when applied at the document level remains unexplored.

\section{Experimental Setup}
Our global aim is to perform a more thorough investigation into the performance of existing document simplification systems, with particular focus on providing more interpretable results that differentiate between faithfulness and simplicity. We also investigate the out-of-domain performance of existing systems and reconsider how this should be evaluated given a lack of diverse references.

\subsection{Data}
We primarily rely on the Newsela~\cite{xu-etal-2015-problems} corpus, which is often considered the gold-standard document simplification dataset. It consists of 1,130 English news articles that have been manually rewritten by professional editors at five different discrete reading levels (0-4) of increasing simplicity.\footnote{We use the same document-level test set as \citet{cripwell-etal-2023-document}.} The main drawback of using Newsela is that it requires a licence to use in research, meaning that it is not necessarily made available to all practitioners. This makes it somewhat more difficult to compare and reproduce results, but unfortunately nothing comes close in terms of quality.

As we intend to focus on reference-less evaluation, we can also consider model performance on out-of-domain data for which we have no reference simplifications. For this, we use standard English Wikipedia (EW) articles from Wiki-auto~\citep{jiang-etal-2020-neural}. Although EW corpora with automatically aligned reference simplifications from simple English Wikipedia (SEW) exist, they are known to contain a lot of noise, being of particularly poor quality when considered at the document level~\citep{xu-etal-2015-problems,cripwell-etal-2023-document}. To assess performance on longer documents, we only consider those that contain at least 10 sentences and 3 paragraphs. To diversify the domain of articles, we annotate each with a semantic type according to their WikiData~\citep{vrandevcic2014wikidata} entry. We select 19 of the most common types, group them into 5 broad categories and sample articles equally from each to obtain a final test set of 1000 documents (further details are given in Appendix~\ref{app:wikidata_annot}).

\subsection{Simplification Systems}
We consider several document simplification systems (at or near state-of-the-art) from existing works, which have all been trained on Newsela.

\textbf{PG$_{\text{Dyn}}$} (Plan-Guided Simplification with Dynamic Context) is a pipeline system that first generates a document simplification plan using high-level context, then conditions a sentence simplification model on said plan~\citep{cripwell-etal-2023-document}. The plan consists of a sequence of simplification operations (split, delete, copy or rephrase), with one for each sentence in the input document. 

From \citet{cripwell-etal-2023-context} we include three additional systems: (i) \textbf{LED$_{\text{para}}$} --- a paragraph-level Longformer~\citep{longformer} model which is the best performing end-to-end system; (ii) \textbf{$\hat{O} \rightarrow$ LED$_{\text{para}}$}, which uses the same Longformer model, but is conditioned on a plan from the same planner as PG$_{\text{Dyn}}$; and (iii) \textbf{$\hat{O} \rightarrow$ ConBART} --- a modification of the BART~\citep{lewis-etal-2020-bart} architecture that attends to a high-level document context during decoding, while also conditioning on a plan.

 Table~\ref{tab:systems} provides a summary of the model attributes.
 
\begin{table}[!htbp]
  \centering
  \small
  \begin{tabular}{p{2.2cm}p{4.5cm}}
  \toprule 
  \bf System & \bf Description \\
  \midrule
  $\text{PG}_{\text{Dyn}}$ & - Sentence-level text input \newline
  - Plan-guided \\
  \midrule
  LED$_{\text{para}}$ & - Paragraph-level text input \newline - No plan-guidance \newline - Longformer-based end-to-end model \\
  \midrule
  $\hat{O} \rightarrow$ LED$_{\text{para}}$ & - Paragraph-level text input \newline - Plan-guided \newline - Longformer-based simplification component \\
  \midrule
  $\hat{O} \rightarrow \text{ConBART}$ & - Sentence-level text input \newline
  - Plan-guided \newline - Simplification model with cross-attention over high-level representation of document sentences \\
  \bottomrule
  \end{tabular}
  \caption{Descriptions of the different document simplification systems we consider.}
  \label{tab:systems}
\end{table}

As these Newsela-trained models have all been prefixed with target reading-level control tokens during training, we must also specify this during inference. For in-domain evaluation, we consider the performance of the various models on each of the four target simplification levels present in Newsela. On the out-of-domain Wikipedia data, we set the target reading-level to 3 (on a scale of 0-4) for all models. Ideally, this will result in substantial editing during simplification while limiting the over-deletion of content. 

\subsection{Evaluating Faithfulness}
We consider two existing reference-less metrics for evaluating faithfulness: \textbf{SummaC} (an NLI entailment-based metric)~\citep{laban-etal-2022-summac} and \textbf{QAFactEval} (a QA-based metric)~\citep{fabbri-etal-2022-qafacteval}. Both are from the summarization literature and should therefore be considered with a level of caution when being applied to simplification. For example, as summarization outputs are generally much shorter than their inputs, it is likely that these metrics will skew in favour of very short and concise simplifications (i.e. precision) even when too much information has been removed. In response, we also use variations of each that focus more on recall.

\paragraph{SummaC (Summary Consistency)}~\citep{laban-etal-2022-summac} first works by using an out-of-the-box NLI model\footnote{In our case, we use an implementation that uses the version of ALBERT-xlarge from \citet{schuster-etal-2021-get} finetuned on the Vitamin C and MNLI datasets, available at \url{https://huggingface.co/tals/albert-xlarge-vitaminc-mnli}.} to compute an NLI entailment matrix over a document. This is an $M \times N$ matrix of entailment scores between each of the $M$ input sentences and $N$ output sentences. This is transformed into a histogram form of each column and a convolutional layer is used to convert the histograms into a single score for each output sentence, which are then averaged. As such, this metric is naturally more precision-oriented and therefore could favour shorter, lexically conservative simplifications. In response, we also compute a recall-oriented version, whereby scores are calculated for each input sentence (i.e. high scores will require generating a simple document that retains as much source information as possible).

\paragraph{QAFactEval}~\citep{fabbri-etal-2022-qafacteval} is a state-of-the-art QA-based metric that consists of several components within a pipeline. In order they are: answer selection $\rightarrow$ question generation $\rightarrow$ question answering $\rightarrow$ overlap evaluation $\rightarrow$ question filtering. Questions and correct answers are first generated given a summary, then answers are predicted given the input document as context. For each of these, an answer overlap score is computed using the LERC metric~\citep{chen-etal-2020-mocha}, which estimates the semantic similarity between the true and predicted answers. The final result is the average of these answer overlap scores for the questions remaining after a question filtering phase (those that are considered answerable). 

If an overly short simplification leads to only a few questions being generated it is possible that this could achieve high scores. Further, the process of simplification itself (lexical subtitution in particular) might challenge this metric as the QA model must be able to accurately recognize the semantic similarity between substituted phrases in order to gauge the validity of an answer. As with SummaC, we compute both precision- and recall-oriented versions of this metric. In the recall case we generate questions from the source document instead of the output.

\paragraph{Entity Matching.} 
Another heuristic for assessing the semantic faithfulness of generated text is to consider the similarity between entities present in the input vs. output --- sometimes referred to as entity-based semantic adequacy (\textbf{ESA})~\citep{wiseman-etal-2017-challenges,laban-etal-2021-keep,faille-etal-2021-entity-based,ma-etal-2022-improving}. We extract named entities from input documents using the spaCy library\footnote{\url{https://spacy.io}} and compute the precision, recall, and F1 with respect to those found in the generated simplifications.

\paragraph{Conservativity.}
Given the nature of semantic faithfulness being tied to the input, high scores for these metrics can be obtained by overly conservative models. So, to better contextualize results, we also include the average lengths of outputs (no. of tokens and sentences) as well as the \textbf{BLEU}~\citep{papineni-etal-2002-bleu} with respect to the input (BLEU$_C$). Generally, simplifications are slightly shorter than their inputs and often contain more sentences (a result of splitting). This BLEU$_C$ score will give a further indication of the amount of editing that has been performed and therefore flag whether a system has potentially achieved high faithfulness scores as a result of over-conservativity.

\subsection{Evaluating Simplicity}
Most popular evaluation metrics for simplification have well documented limitations, such as their reliance on high-quality references. Furthermore, their efficacy has not been fully explored for the document-level task. Given this and the fact that the scope of our study covers performance on out-of-domain data, for which there are no references, we instead rely on reference-less alternatives that are known to correlate well with pure simplicity~\citep{cripwell-etal-2023-simplicity}.

\paragraph{FKGL.} We report the Flesch-Kincaid grade level (FKGL)~\citep{kincaid1975derivation} --- a simple document readability metric with a long history of usage. It is based on a regression model that considers the average length of sentences and syllable count of words in the document. However, FKGL gauges simplicity in absolute terms, assuming a simpler output is universally more valuable. Because of this, it is not ideal for evaluating simplicity for specific target groups (e.g. the different reading grade levels supported by Newsela). 

\paragraph{$\epsilon$SLE$_{doc}$.} Given that most document simplification systems target a specific reading level during generation, it would be more useful to evaluate the divergence from this target level of simplicity, rather than measuring raw simplicity alone. To this end, we modify the \textbf{SLE} sentence level simplicity metric  proposed in ~\citep{cripwell-etal-2023-simplicity} to obtain a simplicity metric for documents which we dub $\epsilon$SLE$_{doc}$.  

\textbf{SLE} is trained to predict a sentence's simplicity level following a leveling scheme similar to Newsela. We adapt this to the document level by computing the prediction for a document $Y$ as the mean of its sentences' \textbf{SLE} scores:

\begin{equation}
  \text{SLE$_{doc}$}(Y) = \frac{1}{\abs{Y}} \sum^{\abs{Y}}_{i=1}{\text{SLE}(y_i)}
\end{equation}

where $y_i$ is the $i$th sentence of document $Y$.
We further adapt this to our task by deriving the simplicity level error (\textbf{$\epsilon$SLE$_{doc}$}) of a system as the mean absolute error (MAE) between the predicted and target document reading levels. 

\begin{equation}
  \epsilon\text{SLE$_{doc}$} = \frac{1}{N} \sum^{N}_{i=1}{\abs{\text{SLE$_{doc}$}(\hat{Y}_i) - l_i}}
\end{equation}

where $l_i$ is a target simplicity level. $\epsilon$SLE is able to estimate how close a simplification is to the target reading level without relying on any references, allowing it to avoid the limitations and rigidity of most other popular evaluation metrics. Although SLE was initially proposed for sentence-level evaluation, it was also trained with document-level labels and to optimize document-level accuracy. As such, we believe SLE$_{doc}$ should work well as a document-level metric. Although individual sentences within a document might have diverse simplicity levels, in aggregate they should converge to the global document level, following the central limit theorem (SLE distributions per reading level are shown in Appendix~\ref{app:sle_dist}).


\section{Results and Discussion}

\subsection{Newsela Performance}
Faithfulness and simplicity results on the Newsela test set are shown in Tables \ref{tab:fact_eval} and \ref{tab:simp_eval_noref}, respectively.

\paragraph{References.} 
We observe that the references  achieve much better FKGL and $\epsilon$SLE$_{doc}$ than any system indicating that simplification models simplify less than Newsela editors. Similarly, all models have higher meaning preservation scores than the references which shows that they are more conservative (since they were hand written by professionals, we assume that references  are sufficiently faithful to the input). This suggests that there is indeed a trade-off between faithfulness and simplicity and more specifically, that models with high meaning preservation scores under-simplify with respect to their target simplification level. 

In summation, there are still improvements that can be made to reduce conservativity and improve simplification in current document simplification systems.



\paragraph{End-to-End vs Planning.} We see a similar trend when comparing end-to-end (LED$_{\text{para}}$) and plan-guided models ($\text{PG}_{\text{Dyn}}$, $\hat{O} \rightarrow$ LED$_{\text{para}}$, $\hat{O} \rightarrow \text{ConBART}$). 
 
The end-to-end model is more meaning preserving than the plan-guided models but simplifies less. 
Specifically, while the end-to-end model (LED$_{\text{para}}$) achieves the highest scores across all three faithfulness metrics, it also has the highest BLEU$_C$, produces outputs that are much longer than the references or any other system and achieves the worst simplicity performance, both in terms of absolute (FKGL) and relative ($\epsilon$SLE$_{doc}$) criteria.

In contrast, the plan-guided models achieve faithfulness results not too far from LED$_{\text{para}}$ while still generating outputs much closer to the references in terms of length and BLEU$_C$. 

Together these results suggest that plan-guidance allows models to avoid conservativity and make necessary edits to achieve high simplicity, although at the cost of some reduced faithfulness to the input. 


\paragraph{Local vs Global Context.} The simplification components of the plan-guided models each consider document context differently. While $\text{PG}_{\text{Dyn}}$ has no notion of document context, $\hat{O} \rightarrow$ LED$_{\text{para}}$ considers the local, token-level context of the surrounding paragraph, and $\hat{O} \rightarrow$ ConBART considers a high-level representation of more global context (SBERT encodings of 26 surrounding sentences). 

The results indicate that the more local paragraph context leads to slight improvement in terms of faithfulness, but a reduction in simplicity performance. $\hat{O} \rightarrow$ ConBART achieves the best overall simplicity (FKGL) as well as $\epsilon$SLE$_{doc}$. Interestingly, both $\hat{O} \rightarrow$ ConBART and $\hat{O} \rightarrow$ LED$_{\text{para}}$ are much better than the other systems at simplifying to the highest level of simplicity (level 4 in Table \ref{tab:fact_eval}), mirroring the human evaluation observations of \citet{cripwell-etal-2023-context} where plan-guided, context-aware systems appeared particularly strong in cases where major editing is required.

\begin{table*}[!htbp]
    \centering
    \small
    \begin{tabular}{lcccccccccccc}
    \toprule
    \bf System & \multicolumn{3}{c}{\bf SummaC $\uparrow$} & \multicolumn{3}{c}{\bf QAFactEval $\uparrow$} & \multicolumn{3}{c}{\bf ESA $\uparrow$} & \multicolumn{2}{c}{\bf Length} & \bf BLEU$_C$ \\
    \midrule
      & P & R & F1 & P & R & F1 & P & R & F1 & Tokens & Sents \\
    \midrule
    Input & - & - & - & - & - & - & - & - & - & 866.9 & 38.6 & - \\
    Reference & 0.61 & 0.47 & 0.53 & 3.86 & 3.02 & 3.39 & 0.59 & 0.47 & 0.52 & 671.5 & 42.6 & 44.6 \\
    \midrule
    $\text{PG}_{\text{Dyn}}$ & 0.65 & 0.47 & 0.55 & 3.95 & 3.10 & 3.47 & \bf0.61 & 0.48 & 0.53 & 667.2 & 42.6 & 47.6 \\
    LED$_{\text{para}}$ & \bf0.66 & \bf0.52 & \bf0.58 & \bf4.00 & \bf3.29 & \bf3.61 & 0.60 & \bf0.51 & \bf0.55 & 712.9 & 44.9 & 51.5 \\
    $\hat{O} \rightarrow$ LED$_{\text{para}}$ & 0.65 & 0.50 & 0.57 & 3.98 & 3.16 & 3.52 & 0.60 & 0.49 & 0.54 & 683.1 & 42.8 & 49.1 \\
    $\hat{O} \rightarrow$ ConBART & 0.65 & 0.48 & 0.56 & 3.95 & 3.11 & 3.48 & 0.60 & 0.48 & 0.53 & 671.6 & 43.0 & 47.5 \\
    \bottomrule
    \end{tabular}
    \caption{\textbf{In-Domain Evaluation.} Faithfulness  results for systems evaluated on the Newsela test set.}
    \label{tab:fact_eval}
\end{table*}

\begin{table*}[!htbp]
    \centering
    \small
    \begin{tabular}{lcccccc}
    \toprule
    \bf System & \bf FKGL $\downarrow$ & \multicolumn{5}{c}{\bf $\epsilon$SLE$_{doc} $$\downarrow$} \\
    \midrule
      &  & 1 & 2 & 3 & 4 & \bf Total \\
    \midrule
    Reference & 4.93 & 0.22 (1.12) & 0.21 (1.97) & 0.24 (3.11) & 0.22 (3.84) & 0.23 \\
    \midrule
    $\text{PG}_{\text{Dyn}}$ & 4.98 & 0.30 (1.24) & \textbf{0.22} (2.02) & 0.22 (3.07) & 0.32 (3.69) & 0.26 \\
    LED$_{\text{para}}$ & 5.15 & 0.29 (1.06) & 0.24 (1.92) & 0.24 (2.97) & 0.34 (3.67) & 0.28 \\
    $\hat{O} \rightarrow$ LED$_{\text{para}}$ & 5.09 & \textbf{0.26} (1.13) & 0.24 (1.87) & 0.23 (3.02) & 0.30 (3.72) & 0.26 \\
    $\hat{O} \rightarrow$ ConBART & \bf4.96 & 0.28 (1.23) & \textbf{0.22} (1.98) & \textbf{0.21} (3.06) & \textbf{0.29} (3.73) & \textbf{0.25} \\
    \bottomrule
    \end{tabular}
    \caption{\textbf{In-Domain Evaluation.} Simplicity results for systems evaluated on the Newsela test set. Columns 1-4 shows the results on the test sets for each level of simplicity, 4 being the level for highest degree of simplification. Numbers in parentheses are the raw SLE averages for each level.  }
    \label{tab:simp_eval_noref}
  \end{table*}

\subsection{Out-of-Domain Performance}
Out-of-domain performance is assessed by testing the Newsela-trained models on EW data. Results are shown in Tables \ref{tab:ood_eval} and \ref{tab:simp_eval_noref_wiki}. The difference in performances between in- and out-of-domain data with the same target reading level is shown in Appendix~\ref{app:extra_results}.

\paragraph{End-to-End vs Planning.} 
The end-to-end, Longformer model (LED$_{\text{para}}$) produces much shorter output documents than the plan-guided models --- the opposite of what is seen for Newsela. 
As EW articles have longer paragraphs on average, this could be a result of over-fitting (i.e. being biased towards Newsela paragraph length observed during training and therefore generating overly short simplifications when applied to the longer EW texts. This could also be a result of over-deletion due to a lack of plan-guidance, as the other paragraph-level model ($\hat{O} \rightarrow$ LED$_{\text{para}}$) does not share this behaviour, potentially suggesting that planning also helps models better adapt to unseen domains.


On the other hand, $\hat{O} \rightarrow$ ConBART achieves the lowest faithfulness scores out of all dedicated systems, particularly on QAFactEval. As this model attends over a wider document context, it is possible that this increase in model variance could have led to some overfitting on the Newsela data. The ConBART network achitecture also contains additional layers that were not pretrained before finetuning on the Newsela dataset, further pointing towards potential overfitting. However, it is still close to $\text{PG}_{\text{Dyn}}$ on SummaC and ESA, while also achieving the best simplicity scores, which could mean the lower faithfulness scores are a result of the trade-off with simplicity. Without reference simplifications, it seems difficult to draw strong conclusions before examining human evaluation results.

\paragraph{Sentences vs Paragraphs.} In terms of simplicity, the sentence-level models ($\text{PG}_{\text{Dyn}}$ and $\hat{O} \rightarrow$ ConBART) achieve much lower FKGL and $\epsilon$SLE$_{doc}$ than the two paragraph-level models. However, like on Newsela, they are markedly outperformed by the paragraph models on faithfulness metrics, particularly in terms of precision. While paragraph models produced longer outputs on in-domain data, they now produce shorter texts than sentence-level models, particularly in terms of the number of sentences. This could indicate potential conservativity with respect to sentence splitting, or an over-deletion of sentences.

\begin{table*}[!htbp]
    \centering
    \small
    \begin{tabular}{lcccccccccccc}
    \toprule
    \bf System & \multicolumn{3}{c}{\bf SummaC $\uparrow$} & \multicolumn{3}{c}{\bf QAFactEval $\uparrow$} & \multicolumn{3}{c}{\bf ESA $\uparrow$} & \multicolumn{2}{c}{\bf Length} & \bf BLEU$_C$ \\
    \midrule
      &  P & R & F1 & P & R & F1 & P & R & F1 & Tokens & Sents \\
    \midrule
    $\text{PG}_{\text{Dyn}}$ & 0.70 & 0.38 & 0.50 & 3.28 & 2.18 & 2.62 & 0.58 & 0.34 & 0.43 & 614.5 & 40.6 & 31.4 \\
    LED$_{\text{para}}$ & \bf0.76 & 0.39 & 0.51 & \bf3.78 & 2.11 & 2.71 & \bf0.64 & 0.35 & 0.45 & 513.7 & 32.5 & 27.4 \\
    $\hat{O} \rightarrow$ LED$_{\text{para}}$ & 0.73 & \bf0.41 & \bf0.53 & 3.61 & \bf2.28 & \bf2.79 & 0.62 & \bf0.37 & \bf0.47 & 601.5 & 37.0 & 32.0 \\
    $\hat{O} \rightarrow$ ConBART & 0.68 & 0.38 & 0.49 & 3.10 & 2.06 & 2.48 & 0.57 & 0.33 & 0.42 & 598.4 & 40.5 & 29.5 \\
    \bottomrule
    \end{tabular}
    \caption{\textbf{OoD Evaluation. }Faithfulness and Conservativity results on the out-of-domain Wikipedia test set.}
    \label{tab:ood_eval}
\end{table*}

\begin{table}[!htbp]
  \centering
  \small
  \begin{tabular}{lcc}
  \toprule
  \bf System & \bf FKGL $\downarrow$ & \bf $\epsilon$SLE$_{doc}$ $\downarrow$ \\
  \midrule
  Input & 10.07 & - (0.89) \\
  \midrule
  $\text{PG}_{\text{Dyn}}$ & 4.72 & \textbf{0.21} (2.92) \\
  LED$_{\text{para}}$ & 4.92 & 0.29 (2.78) \\
  $\hat{O} \rightarrow$ LED$_{\text{para}}$ & 5.02 & 0.31 (2.76) \\
  $\hat{O} \rightarrow$ ConBART & \bf4.58 & \textbf{0.21} (3.00) \\
  \bottomrule
  \end{tabular}
  \caption{\textbf{OoD Evaluation. }Simplicity results on the out-of-domain Wikipedia test set. Numbers in parentheses are the raw SLE$_{doc}$ averages (0-4).}
  \label{tab:simp_eval_noref_wiki}
\end{table}

\section{Human Evaluation}
To confirm system performance on the out-of-domain data, we also conduct a human evaluation. Due to the difficulty of comparing full documents, we follow existing document simplification work in evaluating at the paragraph-level~\cite{cripwell-etal-2023-context}. We present annotators with a complex paragraph and an extract from a generated simplification corresponding to that paragraph. Evaluators are then asked to judge whether the generated text is fluent, consistent with, and simpler than the input.

We randomly sample 250 paragraphs from the test set that contain between 3-6 sentences. We consider the outputs from all tested systems and ask annotators to rate them on each dimension. We pose each as a binary (yes/no) question in order to avoid the inter-annotator subjectivity that is inherent when using a Likert scale. The proportion of positive ratings is used as the final score. Further details are given in Appendix~\ref{app:human_eval}.

\subsection{Human Evaluation Results}
Table~\ref{tab:humeval_results} shows the results of the human evaluation. 

Despite achieving the best fluency, the end-to-end model (LED$_{\text{para}}$) underperforms on both meaning preservation and simplicity compared to the plan-guided systems. This corroborates the automatic results in suggesting that planning can help systems to adapt better to unseen domains. The best overall results are achieved by $\text{PG}_{\text{Dyn}}$, but this can largely be attributed to its very high simplicity ratings as it falls below $\hat{O} \rightarrow$ ConBART in terms of meaning preservation. Although this once again points towards a trade-off between these two dimensions, $\hat{O} \rightarrow$ ConBART manages to achieve the best balance between the two.

In contrast to what is observed via the automatic faithfulness metrics, sentence-level systems also appear to outperform paragraph-level ones. This could be a result of the paragraph models having a wider text window in which to make potential mistakes/hallucinations, whereas the sentence-level systems are more constrained. Further, the EW paragraphs are longer on average than the Newsela ones used to train these models, which could result in them failing to maintain all information when extending to longer input sizes (this is alluded to by the drop in the number of sentences in paragraph-level model outputs when moving to the EW domain, Table~\ref{tab:ood_eval}). In fact, many of the cases where the end-to-end model achieves lower faithfulness scores are the result of the model fully deleting the input paragraphs.


\begin{table}[!htbp]
  \centering
  \small
  \begin{tabular}{lcccc}
  \toprule
  \bf System & Flu & Faith & Simp & \bf Mean \\
  \midrule
  $\text{PG}_{\text{Dyn}}$ & 0.898 & 0.732 & \bf0.820 & \bf0.817 \\
  LED$_{\text{para}}$ & \bf0.932 & 0.632* & 0.664* & 0.743* \\
  $\hat{O} \rightarrow$ LED$_{\text{para}}$ & 0.890 & 0.684 & 0.760 & 0.778* \\
  $\hat{O} \rightarrow$ ConBART & 0.890 & \bf0.760 & 0.764 & 0.805 \\
  \bottomrule
  \end{tabular}
  \caption{Human evaluation results on Wikipedia. Ratings significantly different from the highest rated system on each attribute are denoted with $*$ ($p < 0.05$). Significance was determined with a Student’s $t$-test.}
  \label{tab:humeval_results}
\end{table}

\section{Conclusion}
In this work, we conducted an investigation into the simplicity and the semantic adequacy of outputs from state-of-the-art document simplification systems. By leveraging recent advancements in automatic faithfulness evaluation for summarization and the reference-less evaluation of simplification, we were also able to carry out an analysis of simplification performance on out-of-domain data. 

Separately assessing the models' ability to preserve meaning and simplify allowed for a detailed analysis of how these two dimensions vary across models and between evaluation settings (in- vs out-of-domain evaluation). 

While a state-of-the-art end-to-end model appears to achieve the best in-domain faithfulness results, it is also much more conservative than plan-guided systems, generating outputs with low simplicity. Plan-guided systems also appear better at adapting to unseen domains, but we continue to observe a general trade-off between faithfulness and simplicity. Consideration of this trade-off using only automatic metrics is challenging for out-of-domain settings as it is unclear what exactly constitutes a sufficient level of faithfulness without having references to use as a baseline.

Human evaluation results indicate that plan-guided, sentence-level simplification systems produce outputs with the highest meaning preservation when switching domains --- a phenomenon not captured by the automatic faithfulness metrics. This highlights the need for further exploration into automatic methods of faithfulness evaluation for simplification systems. We hope our work motivates future investigations into more thorough simplification evaluation strategies and the development of training methods and architectures that can allow simplification systems to effectively adapt to unseen domains, rather than further optimizing performance on the most popular datasets.

\section{Limitations}
\paragraph{Paragraph-Level Human Evaluation}
Following previous document simplification studies, our human evaluation was performed using only paragraph-level extracts from simplified documents, rather than the entire documents themselves. This was done to limit the complexity of each human evaluation task as full-document annotation would likely be challenging for many workers. Because of this, it is possible that certain long-distance discourse phenomena are not properly considered during the evaluation. For example, important information may be excluded from a specific output paragraph, but may actually be present in a different part of the document. However, given the iterative nature of most systems tested, such cases should be uncommon. This shift in granularity also makes it difficult to compare automatic and human evaluation results as we cannot directly compute correlations between them. 

\paragraph{English Only}
The datasets and systems we investigate are applicable only to English. It is possible that many of the insights from the study could equally apply in the case of other languages; however, independent analyses would need to be carried out to confirm this. Additionally, many of the evaluation metrics used (e.g. both simplification metrics -- FKGL and SLE) are built specifically with English text in mind and therefore would not easily be adaptable to equivalent evaluations of simplification in other languages.

\section{Acknowledgements}
We thank the anonymous reviewers for their feedback and gratefully acknowledge the support of the Agence Nationale de la recherche, of the Region Grand Est, and Facebook AI Research Paris (Gardent; xNLG, Multi-source, Multi-lingual Natural Language Generation. Award number 199495).

Experiments presented in this paper were carried out using the Grid’5000 testbed, supported by a scientific interest group hosted by Inria and including CNRS, RENATER and several Universities as well as other organizations (see \url{https://www.grid5000.fr}).


\section{Bibliographical References}\label{sec:reference}
\bibliographystyle{lrec-coling2024-natbib}
\bibliography{anthology,custom}

\appendix

\section{WikiData Article Annotation}
\label{app:wikidata_annot}
We selected Wikipedia articles to cover a range of diverse categories (shown in Table~\ref{tab:wiki_categories}). However, we did not obverse any major performance differences between categories, apart from slightly lower scores for articles from more specialized categories (e.g. Science and Industry). 

\begin{table}[!htbp]
  \centering
  \small
  \begin{tabular}{clc}
  \toprule
  \bf Category & \bf Sub-Category & \bf Count \\
  \midrule
  \multirow{ 3}{*}{Biographical} & Human & 500 \\
  & Musical Group & 250 \\
  & Fictional Human & 250 \\
  \midrule
  \multirow{4}{*}{Location} & City & 250 \\
  & Village & 250 \\
  & Commune of France & 250 \\
  & City in the United States & 250 \\
  \midrule
  \multirow{4}{*}{Media} & Film & 250 \\
  & Video Game & 250 \\
  & Literary Work & 250 \\
  & Television Series & 250 \\
  \midrule
  \multirow{4}{*}{Science} & Taxon & 250 \\
  & Class of Disease & 250 \\
  & Chemical Compound & 250 \\
  & Class of Anatomical Entity & 250 \\
  \midrule
  \multirow{4}{*}{Industry} & Business & 250 \\
  & Profession & 250 \\
  & Organization & 250 \\
  & Automobile Model & 250 \\
  \bottomrule
  \end{tabular}
  \caption{Distribution of Wikipedia article categories.}
  \label{tab:wiki_categories}
\end{table}

\section{SLE In-Group Distributions}
\label{app:sle_dist}
\begin{figure}[!htbp]
  \centering
  \includegraphics[width=0.5\textwidth]{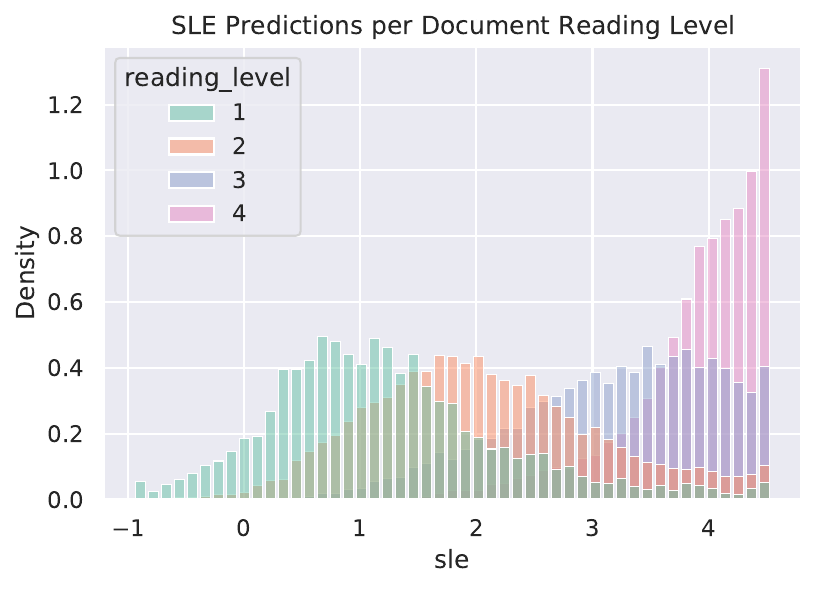}
  \caption{Distribution of SLE scores for reference sentences within each Newsela reading level group.}
  \label{fig:sle_dist_hist}
\end{figure}

Figure~\ref{fig:sle_dist_hist} shows the distribution of SLE scores predicted for reference sentences belonging to each original Newsela reading level group. We can see that although the mean is approximately equal to the reading level, there is substantial diversity within each group.

\section{Human Evaluation Details}
\label{app:human_eval}
\begin{figure*}[!htbp]
  \centering
  \includegraphics[width=0.95\textwidth]{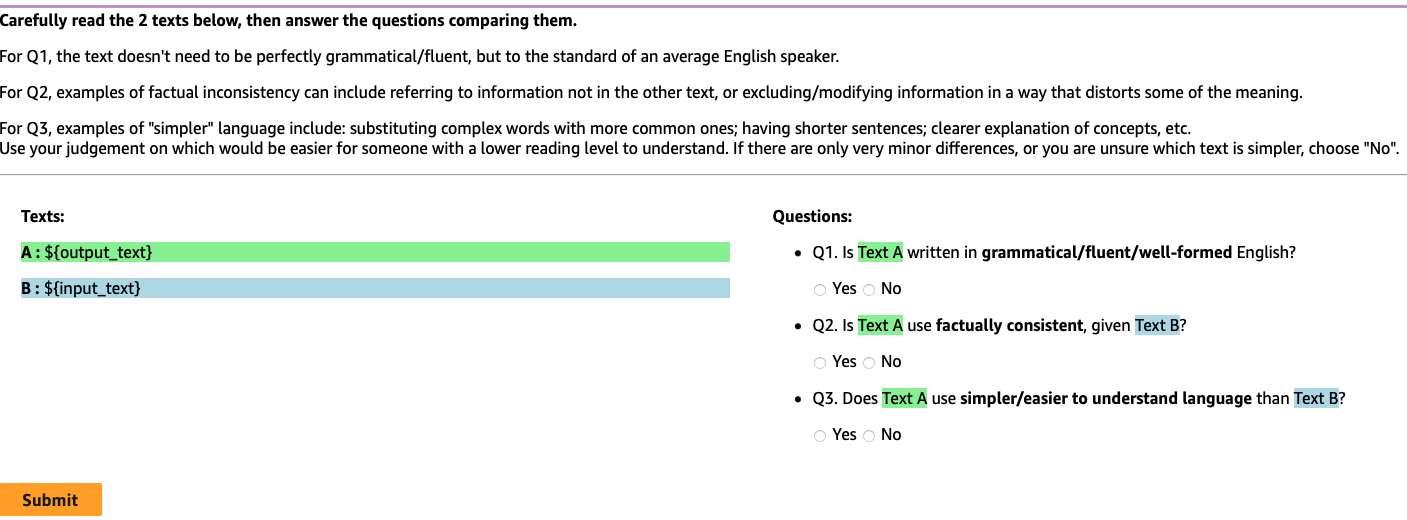}
  \caption{Submission form presented to annotators during the human evaluation.}
  \label{fig:human_eval_form}
\end{figure*}

Human judgements were obtained via the Amazon Mechanical Turk crowd-sourcing platform. Annotators were sourced from majority English speaking countries (AU, CA, GB, IE, NZ, US) and were paid \$0.18 USD per evaluation. According to preliminary tests, under this scheme participants earn approximately \$16.2 USD per hour --- which is higher than the minimum hourly wage of all countries. The form and instructions presented to human evaluators is shown in Figure~\ref{fig:human_eval_form}.

\section{Extra Evaluation Results}
\label{app:extra_results}
Table~\ref{tab:ood_eval_delta} shows the relative change in automatic evaluation results when moving from in- to out-of-domain data (using the same target reading level of 3).

\begin{table*}[!htbp]
  \centering
  \small
  \begin{tabular}{lccccccccc}
  \toprule
  \bf System & \multicolumn{2}{c}{\bf SummaC $\uparrow$} & \multicolumn{2}{c}{\bf QAFactEval $\uparrow$} & \multicolumn{2}{c}{\bf ESA $\uparrow$} & \bf BLEU$_C$ & \bf FKGL $\downarrow$ & \bf $\epsilon$SLE$_{doc}$ $\downarrow$ \\
  \midrule
    &  P & R & P & R & P & R \\
  \midrule
  $\text{PG}_{\text{Dyn}}$ & 0.04 & -0.11 & -0.66 & -0.89 & -0.02 & -0.13 & -14.09 & -0.11 & 0.17 (-0.25) \\
  LED$_{\text{para}}$ & 0.09 & -0.13 & -0.19 & 0.05 & -0.15 & -0.09 & -21.0 & -0.09 & 0.22 (-0.28) \\
  $\hat{O} \rightarrow$ LED$_{\text{para}}$ & 0.07 & -0.1 & -0.33 & -0.84 & 0.03 & -0.11 & -14.55 & 0.06 & 0.24 (-0.32) \\
  $\hat{O} \rightarrow$ ConBART & 0.02 & -0.11 & -0.86 & -1.01 & -0.03 & -0.15 & -16.04 & -0.27 & 0.09 (-0.14) \\
  \bottomrule
  \end{tabular}
  \caption{Difference in results for target-level 3 when moving from the in-domain Newsela to the out-of-domain Wikipedia test set.}
  \label{tab:ood_eval_delta}
\end{table*}

\end{document}